\pdfoutput=1
\documentclass[twocolumn]{fairmeta}

\PassOptionsToPackage{xcdraw,table}{xcolor}

\usepackage{nicematrix}
\usepackage{latexsym}
\usepackage[T1]{fontenc}
\usepackage{amsmath}
\usepackage{amsfonts}
\usepackage{microtype}
\usepackage{inconsolata}
\usepackage{graphicx}
\usepackage{booktabs} %
\usepackage{multirow}
\usepackage{xspace}
\usepackage{subcaption}
\usepackage{url}
\usepackage{comment}
\usepackage[export]{adjustbox}
\usepackage[nameinlink,capitalise]{cleveref}

\usepackage{listings}
\usepackage{xcolor}  %

\usepackage{tabularx}
\usepackage[table]{xcolor}
\usepackage{booktabs}
\usepackage{hyperref}
\definecolor{rowgray}{gray}{0.965}
\newcolumntype{Y}{>{\raggedright\arraybackslash}X}

\usepackage{dblfloatfix} 

\crefformat{section}{§#2#1#3}

\lstdefinelanguage{Prompt}{
  moredelim=[s][\bfseries]{\{\{}{\}\}}, %
}

\newcommand{\translateClassify}{\textsc{\texttt{translate-classify}}\xspace}
\newcommand{\classify}{\textsc{\texttt{classify}}\xspace}
\newcommand{\classifyID}{\textsc{\texttt{classify (ID)}}\xspace}
\newcommand{\classifyOOD}{\textsc{\texttt{classify (OOD)}}\xspace}
\newcommand{\translateJudge}{\textsc{\texttt{translate-judge}}\xspace}
\newcommand{\judge}{\textsc{\texttt{judge}}\xspace}
\newcommand{\translateClassifyLlama}{\textsc{\texttt{translate-classify (Llama 3)}}\xspace}
\newcommand{\translateClassifyTowerBlocks}{\textsc{\texttt{translate-classify (+TowerBlocks/MT)}}\xspace}

\title{Translate, then Detect: Leveraging Machine Translation for Cross-Lingual Toxicity Classification} 

\author[1,*]{Samuel J. Bell}
\author[1,2,*]{Eduardo Sánchez}
\author[1]{David Dale}
\author[2]{Pontus Stenetorp}
\author[3]{Mikel Artetxe}
\author[1]{Marta R. Costa-jussà}

\affiliation[1]{FAIR at Meta}
\affiliation[2]{University College London}
\affiliation[3]{University of the Basque Country (UPV/EHU)}

\contribution[*]{Joint first author}

\date{\today}
\correspondence{\email{sjbell@meta.com}, \email{eduardosanchez@meta.com}}

\abstract{Multilingual toxicity detection remains a significant challenge due to the scarcity of training data and resources for many languages.
While prior work has leveraged the \textit{translate-test} paradigm to support cross-lingual transfer across a range of classification tasks, the utility of translation in supporting toxicity detection at scale remains unclear.
In this work, we conduct a comprehensive comparison of translation-based and language-specific/multilingual classification pipelines.
We find that translation-based pipelines consistently outperform out-of-distribution classifiers in 81.3\% of cases (13 of 16 languages), with translation benefits strongly correlated with both the resource level of the target language and the quality of the machine translation (MT) system.
Our analysis reveals that traditional classifiers outperform large language model (LLM) judges, with this advantage being particularly pronounced for low-resource languages, where \texttt{translate-classify} methods dominate \texttt{translate-judge} approaches in 6 out of 7 cases.
We additionally show that MT-specific fine-tuning on LLMs yields lower refusal rates compared to standard instruction-tuned models, but it can negatively impact toxicity detection accuracy for low-resource languages.
These findings offer actionable guidance for practitioners developing scalable multilingual content moderation systems.
}

\begin{document}

\maketitle

\section{Introduction}

Detecting instances of toxic, abusive, or hateful content at scale is a challenging problem with important, real-world implications for content moderation. 
In a multilingual setting, however, toxicity detection is often rendered particularly difficult due to a paucity of labeled data for lower-resourced languages.
In parallel, recent years have seen the scaling up of machine translation (MT) systems to cover a vast array of world languages (e.g.,~\citealp{nllbteam2022language}), offering a potential pathway toward leveraging cross-lingual transfer for improved multilingual toxicity detection.

\begin{figure}[t!]
    \centering
    \includegraphics[width=\linewidth,trim={0.7cm 0 0.7cm 0},clip]{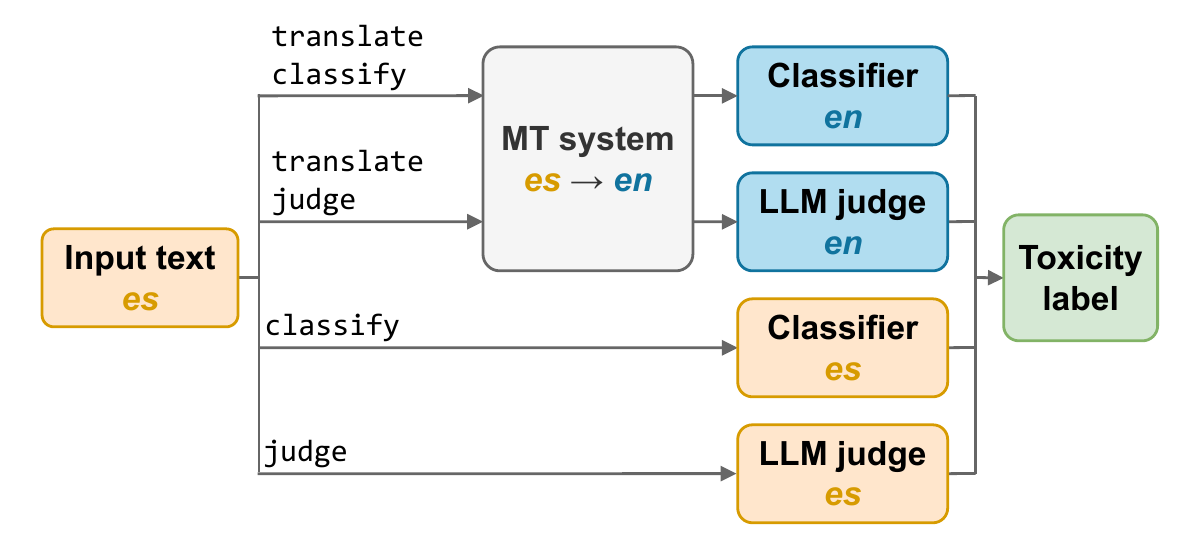}
    \caption{Across 17 languages, we evaluate toxicity detection using translation-based pipelines (\translateClassify, \translateJudge) against classifying in the original language (\classify, \judge). In this example, text in Spanish (es) is optionally translated to English (en) before classification.} 
    \label{fig:diagram}
\end{figure}

In monolingual non-English settings, cross-lingual transfer has already proven useful for toxicity detection \citep{eskelinen-etal-2023-toxicity, 10.1145/3539637.3556892}, aligning with broader analyses of translation's utility for cross-lingual transfer across a range of classification tasks \citep{artetxe-etal-2023-revisiting,etxaniz2023multilinguallanguagemodelsthink,ponti2021modellinglatenttranslationscrosslingual}.
Specifically, \citet{artetxe-etal-2023-revisiting} compare \textit{translate-test} (translating a sample before zero-shot classification) against \textit{translate-train} (translating a sample before classification with a classifier \emph{finetuned on translation data}) and find that \textit{translate-test} is competitive as long as translation quality is sufficient.

While cross-lingual classification has been widely studied in other NLP tasks, toxicity detection presents distinctive challenges that warrant separate investigation.
Toxic language is culturally and contextually grounded, with expressions, slurs, and taboos that often lack direct equivalents across languages, making transfer more brittle than for semantically simpler labels.
Online toxicity also frequently involves code-switching, orthographic variation, and deliberate obfuscation, which may be less common in other tasks.
Moreover, toxicity labels are inherently subjective and shaped by cultural norms, leading to potential label drift when transferring across languages.

These factors, combined with the high stakes of moderation errors, make cross-lingual transfer in toxicity detection both consequential and scientifically challenging.

In this work, we present an empirical exploration of translation for multilingual toxicity detection, through the lens of the practitioner for whom labeled data may be unavailable---a particularly common scenario when working with lower-resourced languages---by comparing translation-based pipelines against a variety of off-the-shelf multi- and monolingual classifiers.
Across 27 pipelines spanning five MT systems and nine toxicity classifiers---including both traditional classifiers and large language model (LLM) judges---we evaluate the benefit of cross-lingual classification in 17 languages with varying levels of resources.

Our results suggest that leveraging translation is an effective method for multilingual toxicity detection (\cref{sec:results-overall}), with benefits scaling in line with increasing language resources and MT system quality (\cref{sec:results-resources}).
Motivated by these results, we study the issue of refusal rates and its mitigation via MT supervised finetuning (MT-SFT), as well as the downstream effect of MT-SFT on toxicity detection performance (\cref{sec:results-finetuned-and-refusal}).
Finally, we explore classifying using LLM judges and compare them to traditional toxicity classifiers (\cref{sec:results-llm-judge}).
We conclude with practical recommendations for deploying multilingual toxicity detection systems at scale.

\section{Related work}
\label{sec:relwork}

\subsection{Multilingual toxicity detection}
Multilingual toxicity detection is widely used in cases like content moderation or faithful translation (e.g.~\citealp{costa-jussa-etal-2023-toxicity}).
Prior work has either trained models using multilingual corpora of labeled training data (e.g.~\citealp{hanu2020detoxify}), or sought to exploit cross-lingual transfer via monolingual finetuning of multilingual foundation models (e.g.~XLM-ROBERTa; \citealp{conneau2020unsupervised}). 
Multilingual evaluation datasets exist for toxicity detection (e.g.~\citealp{kivlichan2020jigsaw,gupta-2021-multilingual}) alongside those used for text detoxification \citep{dementieva2024overview,dementieva-etal-2025-multilingual}.
In this work, we evaluate a representative sample of off-the-shelf traditional classifiers, including cross-lingual, and both mono- and multilingual classifiers, across a wide variety of languages.

\subsection{Cross-lingual classification}

Early approaches to cross-lingual classification relied on bilingual lexicons and statistical methods to project documents into a shared feature space \citep{rapp1995identifying,dumais1997automatic,gliozzo2006exploiting}. The introduction of cross-lingual word embeddings \citep{mikolov2013exploiting,faruqui2014improving,ammar2016massively} enabled models trained in one language to be applied to others through a shared vector space. Prior to multilingual encoders, transfer was typically achieved via MT, either by translating the training data into the target language (\textit{translate-train}) or by translating inputs into the source language at inference (\textit{translate-test}) \citep{wan2009co,prettenhofer2010cross}.

Multilingual sentence encoders such as LASER \citep{artetxe2019massively} and mBERT \citep{devlin2019bert} demonstrated the feasibility of direct zero-shot transfer without translation. XLM \citep{lample2019cross} introduced translation language modeling to improve alignment, and XLM-R \citep{conneau2020xlmr} showed consistent gains from scaling model and data size. \citet{artetxe2020translation} provided a systematic comparison of translate-train and translate-test, while \citet{etxaniz2023translatetest} revisited translate-test with modern neural MT, finding it competitive for low-resource and distant languages.

Recent work explores large multilingual LLMs \citep{muennighoff2022crosslingual} and parameter-efficient adaptation methods \citep{pfeiffer2020madx}, aiming to combine the flexibility of fine-tuning with the scalability of zero-shot prompting.

\section{Methods}
\label{sec:methods}

We evaluate the performance of toxicity detection \emph{pipelines}, where a pipeline comprises a binary toxicity classifier and an optional MT system.
In many languages---and particularly for lower-resourced languages---labeled data for toxicity detection is unavailable, precluding the training and deployment of specialized classifiers and motivating the consideration of translation-based pipelines.
As such, we are principally interested in comparing pipelines in the following three regimes:

\begin{table*}[t]
\footnotesize
\centering
\begin{NiceTabular}{llrlr}
\toprule
Language Code & Language & FineWeb-2 Docs & Dataset & No. Samples \\
\midrule
am & Amharic & 280,355 & \href{https://huggingface.co/datasets/uhhlt/amharichatespeechranlp}{Amharic Hate Speech}~\citep{ayele2023exploring} & 1,501 \\
ar & Arabic & 57,752,149 & \href{https://github.com/Hala-Mulki/L-HSAB-First-Arabic-Levantine-HateSpeech-Dataset}{L-HSAB}~\cite{mulki2019lhsab} & 5,846 \\
de & German & 427,700,394 & \href{https://github.com/uds-lsv/GermEval-2018-Data}{GermEval 2018}~\cite{wiegand2018overview} & 3,398 \\
es & Spanish & 405,634,303 & \href{https://www.kaggle.com/competitions/jigsaw-multilingual-toxic-comment-classification}{Jigsaw Multilingual}~\cite{kivlichan2020jigsaw} & 8,438 \\
fr & French & 332,646,715 & \href{https://www.kaggle.com/competitions/jigsaw-multilingual-toxic-comment-classification}{Jigsaw Multilingual}~\cite{kivlichan2020jigsaw} & 10,920 \\
he & Hebrew & 13,639,095 & \href{https://github.com/SinaLab/OffensiveHebrew}{OffensiveHebrew}~\cite{hamad2023offensive} & 500 \\
hi & Hindi & 20,587,135 & \href{https://github.com/ShareChatAI/MACD}{MACD}~\cite{gupta2022multilingual} & 6,728 \\
it & Italian & 219,117,921 & \href{https://www.kaggle.com/competitions/jigsaw-multilingual-toxic-comment-classification}{Jigsaw Multilingual}~\cite{kivlichan2020jigsaw} & 8,494 \\
kn & Kannada & 2,309,261 & \href{https://github.com/ShareChatAI/MACD}{MACD}~\cite{gupta2022multilingual} & 6,587 \\
ml & Malayalam & 3,406,035 & \href{https://github.com/ShareChatAI/MACD}{MACD}~\cite{gupta2022multilingual} & 5,170 \\
pt & Portuguese & 189,851,449 & \href{https://huggingface.co/datasets/mteb/told-br}{ToLD-Br}~\cite{leite2020toxic} & 21,000 \\
ru & Russian & 605,468,615 & \href{https://www.kaggle.com/datasets/blackmoon/russian-language-toxic-comments}{Russian Language Toxic Comments}~\cite{belchikov2019russian} & 14,412 \\
ta & Tamil & 5,450,192 & \href{https://github.com/ShareChatAI/MACD}{MACD}~\cite{gupta2022multilingual} & 6,000 \\
te & Telugu & 2,811,760 & \href{https://github.com/ShareChatAI/MACD}{MACD}~\cite{gupta2022multilingual} & 6,000 \\
th & Thai & 35,949,449 & \href{https://github.com/tmu-nlp/ThaiToxicityTweetCorpus}{Thai Toxicity Tweet Corpus}~\cite{sirihattasak2018annotation} & 2,794 \\
tr & Turkish & 88,769,907 & \href{https://www.kaggle.com/competitions/jigsaw-multilingual-toxic-comment-classification}{Jigsaw Multilingual}~\cite{kivlichan2020jigsaw} & 14,000 \\
uk & Ukrainian & 47,552,562 & \href{https://huggingface.co/datasets/textdetox/multilingual_toxicity_dataset}{TextDetox 2024}~\cite{dementieva2024toxicity} & 5,000 \\
\bottomrule
\end{NiceTabular}
\caption{Toxicity datasets used per language, including number of samples, and number of documents in FineWeb-2 as a measure of language resourcedness.}
\label{table:datasets}
\end{table*}

\textbf{\classifyID} \quad An untranslated, in-distribution (ID) sample is classified in the source language using a classifier trained on data from the same distribution (e.g., evaluating a classifier on French social media posts that has been trained on French social media posts).

\textbf{\classifyOOD} \quad An untranslated, out-of-distribution (OOD) sample is classified in the source language using a classifier trained on data from a different distribution (e.g., evaluating a classifier on French video comments that has been trained on French social media posts).

\textbf{\translateClassify} \quad The sample is translated into English using an MT model before being classified in English, using a toxicity classifier that supports English. No evaluated classifiers have been trained on translated data. 
\\

While we expect finetuned classifiers to exhibit the strongest performance while operating ID, it is relative to the far more common OOD scenario (i.e., where no suitably finetuned classifier is available to process the source language) that we expect \texttt{translate} pipelines to offer significant utility.

\subsection{Evaluation}
\label{sec:methods-evaluation}

We evaluate various pipeline implementations across several languages and datasets, each of which comprises text samples $x_i$ and gold toxicity labels $y_i$.
Each pipeline, given a sample, produces a continuous score corresponding to toxicity.

\textbf{Pipeline performance} \quad
To avoid the need for thresholding, we evaluate pipeline performance via the Area Under the Receiver Operating Characteristic curve (AUC), which provides a continuous measure of how well the pipeline can separate toxic from non-toxic samples. 
The AUC is defined as:
\begin{align*}
\text{AUC} = \int_{0}^{1} \text{TPR}(t) \, d\text{FPR}(t)
\end{align*}
where $\text{TPR}(t)$ and $\text{FPR}(t)$ are the true positive and false positive rates at threshold \(t\).
 
When comparing pipelines, we typically evaluate the benefit of using one pipeline over another by way of change in AUC.
For two pipelines, $P_A$ and $P_B$,
\begin{align*}
    \Delta \textrm{AUC}(P_A, P_B) = \textrm{AUC}(P_A) - \textrm{AUC}(P_B)
\end{align*}

We evaluate all possible combinations of pipeline and dataset where the supported pipeline language matches the dataset's language.

\begin{table*}[t]
\centering
\small
\setlength{\tabcolsep}{5pt}
\renewcommand{\arraystretch}{1.12}
\rowcolors{2}{white}{rowgray}
\begin{tabularx}{\textwidth}{Y p{3.5cm} l Y}
\toprule
Classifier & Supported Languages & Base Model & Training Dataset \\
\midrule
\href{https://huggingface.co/Newtral/xlm-r-finetuned-toxic-political-tweets-es}{xlm-r-finetuned-toxic-political-tweets-es} & es & XLM-RoBERTa & Tweets by Spanish politicians \\
\href{https://huggingface.co/citizenlab/distilbert-base-multilingual-cased-toxicity}{distilbert-base-multilingual-cased-toxicity} & 102 languages & DistilBERT multilingual & Jigsaw \\
\href{https://huggingface.co/ml6team/distilbert-base-german-cased-toxic-comments}{distilbert-base-german-cased-toxic-comments} & de & German DistilBERT & Various incl. GermEval 2018 \\
\href{https://huggingface.co/s-nlp/russian_toxicity_classifier}{russian\_toxicity\_classifier} \citep{dementieva2022russe2022} & ru & RuBERT & Russian Language Toxic Comments \\
\href{https://huggingface.co/textdetox/xlmr-large-toxicity-classifier}{xlmr-large-toxicity-classifier} & am, ar, de, en, es, hi, ru, uk, zh & XLM-RoBERTa & TextDetox 2024 \citep{dementieva2024overview} \\
\href{https://huggingface.co/uhhlt/amharic-hate-speech}{amharic-hate-speech} & am & Amharic RoBERTa & Amharic Hate Speech \\
\href{https://huggingface.co/unitary/multilingual-toxic-xlm-roberta}{multilingual-toxic-xlm-roberta} \citep{hanu2020detoxify} & en, es, fr, it, pt, ru, tr & XLM-RoBERTa & Jigsaw Multilingual \\
\href{https://huggingface.co/unitary/toxic-bert}{toxic-bert} \citep{hanu2020detoxify} & en & BERT & Jigsaw \\
\bottomrule
\end{tabularx}
\caption{Open-source toxicity classifiers evaluated in this work.}
\label{table:models}
\end{table*}

\begin{table*}[t]
\centering
\small
\begin{tabular}{ll}
\toprule
Model & Type \\
\midrule
\href{https://huggingface.co/meta-llama/Llama-3.1-8B-Instruct}{Llama 3.1 8B Instruct} \cite{grattafiori2024llama} & LLM \\
\href{https://huggingface.co/google/gemma-3-4b-it}{Gemma 3 4B Instruct} \cite{gemmateam2025gemma} & LLM \\
GPT-4o \cite{openai2024gpt4o} & LLM \\
\href{https://huggingface.co/facebook/nllb-200-3.3B}{NLLB 200 3.3B} \cite{nllbteam2022language} & NMT \\
\bottomrule
\end{tabular}
\caption{Translation systems evaluated in this work.}
\label{table:translation-systems}
\end{table*}

\textbf{Language resources} \quad
We evaluate the role of language resourcefulness on pipeline performance, where we roughly approximate the number of available resources using the amount of documents available in FineWeb2 \citep{penedo2025fineweb2}, a large-scale dataset of web text sourced from various CommonCrawl snapshots.

\textbf{Translation system quality} \quad
Following standard practice (e.g.,~\citealt{kocmi2024findings}), we additionally evaluate the quality of translations into English using the COMETKiwi-DA-XL (\citealt{rei2023scaling}) quality estimation model, evaluated on the BOUQuET \citep{omnilingualmtteam2025bouquet} dataset.

\subsection{Datasets}
\label{sec:methods-data}

We curate a set of ten toxicity benchmarks for evaluating pipeline performance, spanning 17 languages, where each dataset comprises samples of text with gold labels indicating toxicity. 
Benchmarks were identified via searching related work on toxicity detection and by searching the Hugging Face datasets catalog.
We limited our search to only datasets comprising natural human data, and to those where the gold labels are produced by human annotators, such that datasets comprising model-generated or otherwise synthetic text or labels were discarded.
Datasets were restricted to those with a permissive license, where data provenance was clearly indicated, and where the data is readily-accessible online. 
This resulted in the following benchmarks:
Amharic Hate Speech \citep{ayele2023exploring}; GermEval 2018 (German; \citealt{wiegand2018overview}); Jigsaw Multilingual (Spanish, French, Italian, and Turkish partitions only; \citealt{kivlichan2020jigsaw}); L-HSAB (Levantine Arabic; \citealt{mulki2019lhsab}); MACD (Hindi, Kannada, Malayalam, Tamil, and Telugu; \citealt{gupta2022multilingual}); OffensiveHebrew \cite{hamad2023offensive}; ToLD-Br (Brazilian Portuguese; \citealt{leite2020toxic}); Russian Language Toxic Comments \cite{belchikov2019russian}; Thai Toxicity Tweet Corpus \cite{sirihattasak2018annotation}; and TextDetox 2024 (Ukrainian partition only; \citealt{dementieva2024toxicity}).
See \cref{table:datasets} for full details.

Across all datasets, only the test partition is used for evaluation. 
Where a toxicity classifier is trained on data that includes one of our benchmark’s training partitions, we consider that classifier to be operating ID.
Otherwise, as the classifier has been trained on data unlike the benchmark, we consider it to be operating OOD. 
See \cref{table:models} for the training data used to produce each classifier.

\begin{figure*}[!th]
    \centering
    \includegraphics[width=\linewidth]{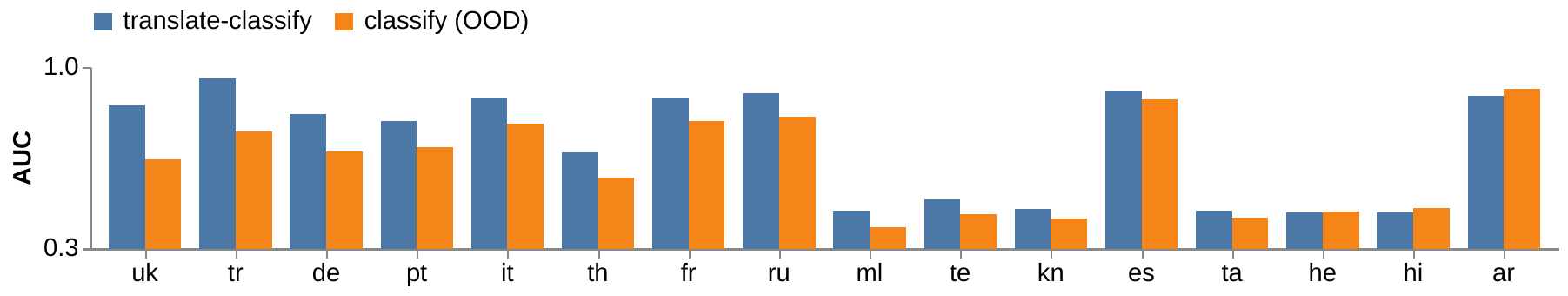}
    \caption{AUC of best possible \translateClassify pipeline (over all combinations of translation systems and English toxicity classifiers) and best possible \classifyOOD pipeline (over all OOD toxicity classifiers). \textbf{The \translateClassify approach wins across 13 out of 16 evaluated languages.}}
    \label{fig:absolute-performance-best-possible-vs-ood}
\end{figure*}

For the purposes of our evaluation, we intentionally avoid distinguishing between toxicity detection and and hate speech detection.
While hate speech and toxic or offensive are distinct concepts \citep{davidson2017automated,waseem2017understanding}---with hate speech typically being interpreted as directed toward a specific group \citep{davidson2017automated,rottger2021hatecheck}---in practice, most evaluation datasets use the terms toxicity, abusive or offensive language, and hate speech almost interchangeably \citep{fortuna2020toxic,banko2020unified}.
As a result we consider datasets spanning toxicity and hate speech detection, and expect minimal difference in findings between tasks.

\subsection{Toxicity classifiers}
\label{sec:methods-classifiers}

We consider eight open-source toxicity classifiers, including English-language, non-English monolingual, and multilingual, all of which are available on Hugging Face.
See \cref{sec:app-toxicity-classifier-selection} for selection criteria and \cref{table:models} for full details of all classifiers considered.

All classifiers evaluated make use of pretrained Transformer-based encoder models as a backbone, such as BERT \citep{devlin2019bert}, DistilBERT \citep{sanh2020distilbert}, ROBERTa \citep{liu2019roberta}, or the multilingual XLM-ROBERTa \citep{conneau2020unsupervised}, some of which have undergone additional fine-tuning on language specific corpora, such as Russian RuBERT \citep{kuratov2019adaptation}.
All classifiers are then finetuned on a portion of labeled toxicity data, such as detailed in \cref{sec:methods-data}.

\subsection{Translation systems}

For \translateClassify pipelines, we translate samples into English with a translation system before classifying the translations with an English-supporting classifier.
We evaluate four different translation systems (see \cref{table:translation-systems}), including both encoder-decoder MT systems (NMT) and decoder-only (i.e., LLM) translation systems.
In the NMT category, we use NLLB 200 3.3B \citep{nllbteam2022language}.
We evaluate three LLM systems (two open-weights and one behind-API): Llama 3.1 8B Instruct \citep{grattafiori2024llama}, Gemma 3 4B Instruct \citep{gemmateam2025gemma} and GPT-4o \citep{openai2024gpt4o}. The following prompt is used to produce the translations: \hfill \newline
\begin{lstlisting}
Translate the following sentence from  {{lang}} into English. Respond only with the translation into English, without any additional comments.
{{sentence}}
\end{lstlisting}

\section{Experiments}
\label{sec:results}

\subsection{Translated pipelines often win}
\label{sec:results-overall}

We compare the AUC of the best \translateClassify pipeline (the best possible combination of translation system and toxicity classifier) against the best possible \classify pipeline (the best toxicity classifier that supports each language).

\begin{figure}[t]
    \centering
    \includegraphics[width=\linewidth]{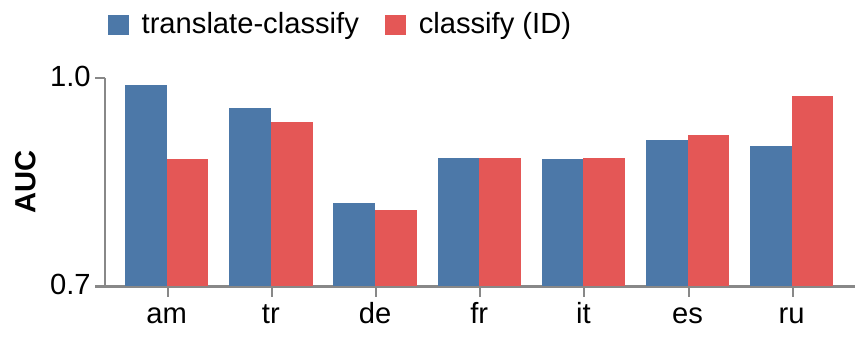}
    \caption{AUC of best possible \translateClassify pipeline (over all combinations of translation systems and English toxicity classifiers) and best possible \classifyID pipeline (over all ID toxicity classifiers).  \textbf{The \translateClassify approach still wins across three of seven languages where in-distribution finetuned classifiers are available.}}
    \label{fig:absolute-performance-best-possible-vs-id}
\end{figure}

\textbf{Results} \quad
In \cref{fig:absolute-performance-best-possible-vs-ood}, we evaluate \translateClassify in the common scenario where a language-specific finetuned toxicity classifier is unavailable, i.e., where classifiers are operating OOD with respect to either their source language or training domain, \classifyOOD.
We observe that in such a scenario, the best \translateClassify pipeline outperforms the best \classifyOOD pipeline across 13 of 16 languages considered (81.3\%).
Reducing a degree of freedom by using a fixed classifier, \texttt{distilbert-base-multilingual-cased -toxicity}, \translateClassify still outperforms \classify in 12 of 16 languages (75\%; see \cref{fig:absolute-performance-best-possible-fixed-classifier-vs-ood}).

\begin{figure*}[!t]
    \centering
    \hfill
    \begin{subfigure}[t]{0.245\textwidth}
        \centering 
        \includegraphics[height=3.55cm]{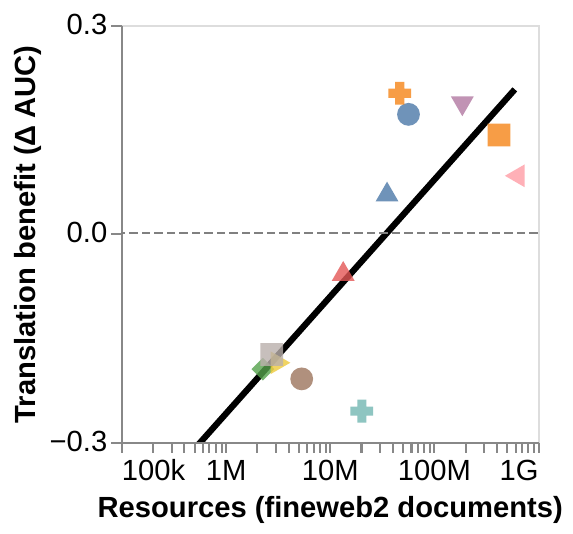}
        \caption{GPT-4o}
        \label{fig:translation-benefit-vs-resources-gpt4o}
    \end{subfigure}
    \hfill
    \begin{subfigure}[t]{0.245\textwidth}
        \centering 
        \includegraphics[height=3.55cm]{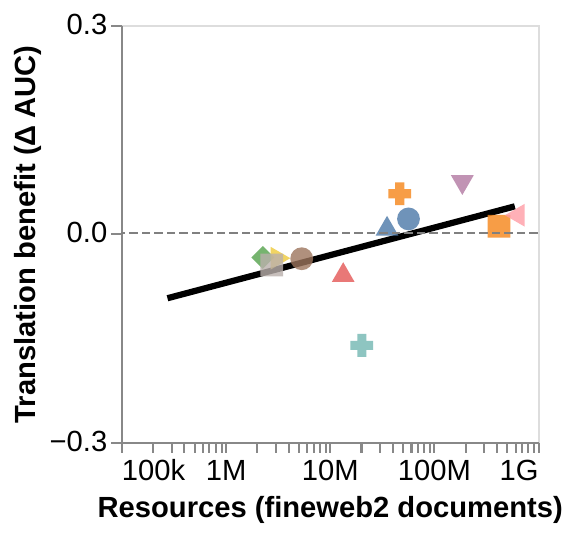}
        \caption{Llama 3.1 8B Instruct}
        \label{fig:translation-benefit-vs-resources-llama-3}
    \end{subfigure}
    \hfill
    \begin{subfigure}[t]{0.245\textwidth}
        \centering 
        \includegraphics[height=3.55cm]{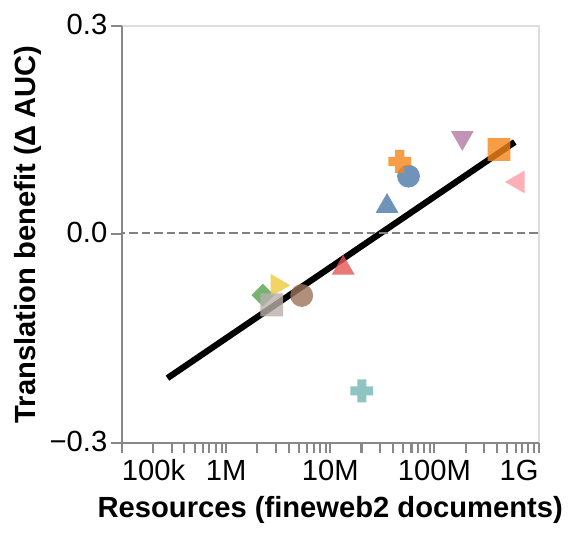}
        \caption{Gemma 3 4B Instruct}
        \label{fig:translation-benefit-vs-resources-gemma-3}
    \end{subfigure}
    \hfill
    \begin{subfigure}[t]{0.245\textwidth}
        \centering 
        \includegraphics[height=3.55cm]{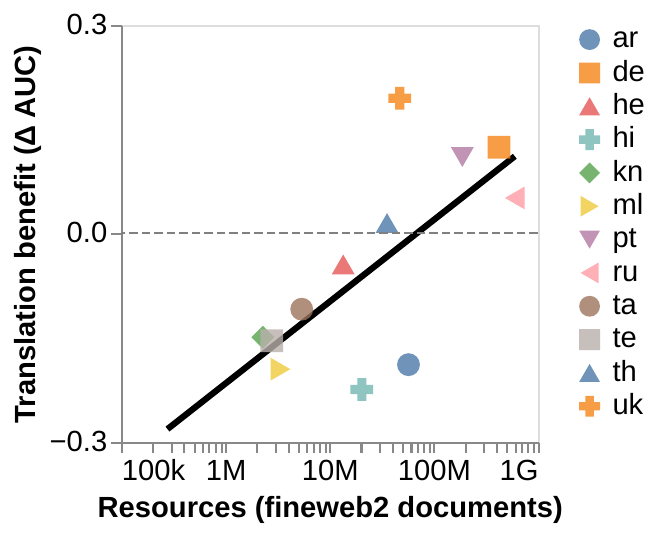}
        \caption{NLLB}
        \label{fig:translation-benefit-vs-resources-nllb}
    \end{subfigure}
    \hfill
    \caption{Change in AUC (i.e., translation benefit) between \translateClassify pipelines with a fixed English classifier, \texttt{toxic-bert}, and \classify pipelines with a fixed multilingual classifier, \texttt{distilbert-base-multilingual-cased-toxicity}, as a function of language resources, over four translation systems \textbf{(a)} GPT-4o, \textbf{(b)} Llama 3.1 8B Instruct, \textbf{(c)} Gemma 3 4B Instruct, and \textbf{(d)} NLLB. \textbf{Translation benefit is increased for higher resourced languages.}}
    \label{fig:translation-benefit-vs-resources}
\end{figure*}

\begin{figure*}[!t]
    \centering
    \hfill
    \begin{subfigure}[t]{0.45\linewidth}
        \centering 
        \includegraphics[height=3.55cm]{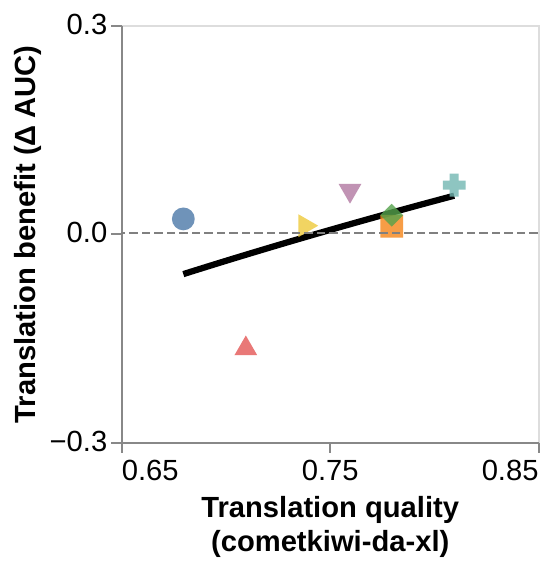}
        \caption{Llama 3.1 8B Instruct}
        \label{fig:translation-benefit-vs-quality-llama-3}
    \end{subfigure}
    \hspace{4pt}
    \begin{subfigure}[t]{0.45\linewidth}
        \centering 
        \includegraphics[height=3.55cm]{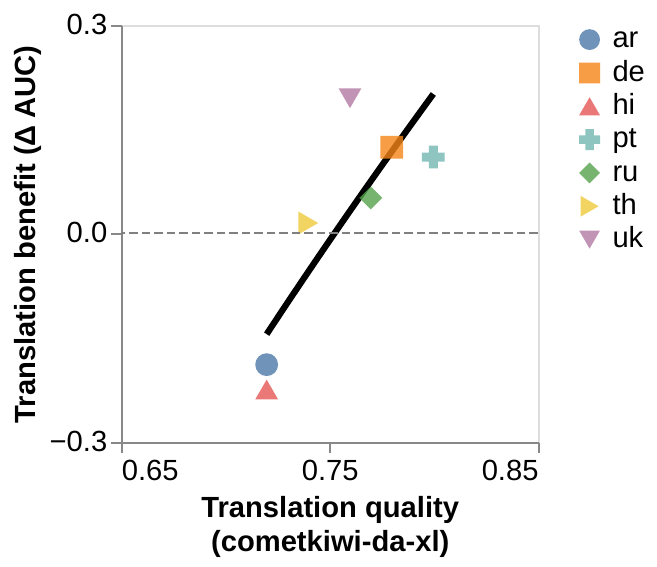}
        \caption{NLLB}
        \label{fig:translation-benefit-vs-quality-nllb}
    \end{subfigure}
    \hfill
    \caption{Change in AUC (i.e., translation benefit) between \translateClassify pipelines and \classify pipelines, as a function of English translation quality measured by \texttt{cometkiwi-da-xl}, over two translation systems \textbf{(a)} Llama 3.1 8B Instruct, and \textbf{(b)} NLLB. \textbf{Translation benefit increases with translation quality for both LLM-based and NMT systems.}}
    \label{fig:translation-benefit-vs-quality}
    \hfill
\end{figure*}

In \cref{fig:absolute-performance-best-possible-vs-id} we evaluate translated pipelines in scenarios where a language-specific finetuned classifier 
\emph{is} available (\classifyID), though we note that this is far from the case for the majority of languages. 
Here, \translateClassify still offers a robust baseline, outperforming finetuned \classifyID pipelines across three out of seven languages.
See \cref{table:overall-results} for full results over all languages.

\subsection{Translation benefit scales with resources}
\label{sec:results-resources}

Next, we explore which factors determine the success of \translateClassify pipelines.
To allow for consistent comparison across languages and control for variability in classifier performance, we now limit ourselves to two fixed classifiers: for \translateClassify we use the English classifier, \texttt{toxic-bert}, while for \classify we use our most multilingual classifier, \texttt{distilbert-base-multilingual-cased -toxicity}.
We evaluate the role of language resourcefulness and translation quality on change in AUC between pipelines, as specified in \cref{sec:methods-evaluation}.

\textbf{Results} \quad
In \cref{fig:translation-benefit-vs-resources}, we observe that the relative benefit of \translateClassify over \classify, as measured by the change in AUC, is higher for better-resourced languages.
This is consistent across four different translation systems, including both LLM and NMT systems. After fixing the best performing classifiers, we notice that the relative benefit of translation for some languages is affected, suggesting the framework is susceptible to model selection to maximize gains.

Similarly, in \cref{fig:translation-benefit-vs-quality} we see that the relative benefit of \translateClassify increases with the quality of translations in each language, across both LLM and NMT systems. 
We note a higher sensitivity to both language resourcefulness and translation quality for the NMT system, NLLB, compared with LLM systems.

\subsection{MT-SFT reduces refusal and improves performance}
\label{sec:results-finetuned-and-refusal}

\begin{figure*}[!th]
    \centering
    \includegraphics[width=\linewidth,trim={0.25cm 0 0.25cm 0},clip]{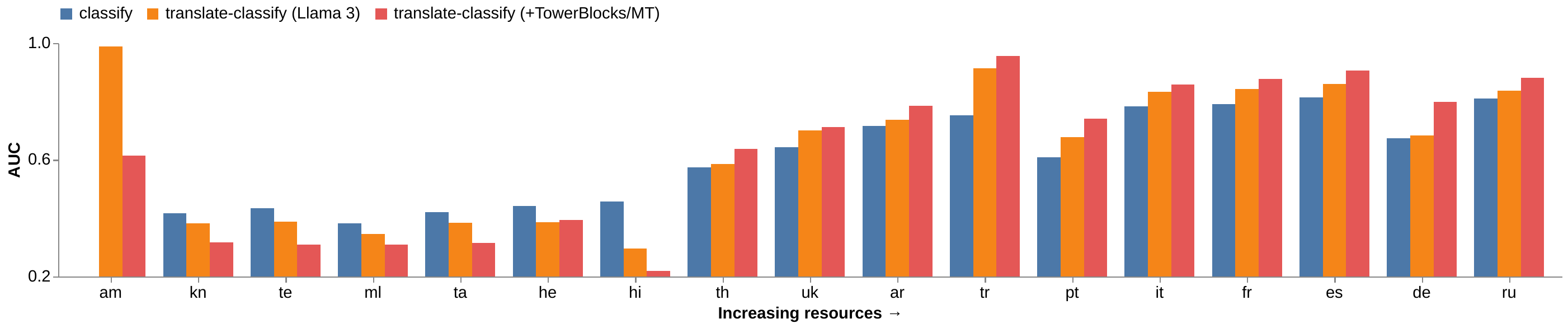}
    \caption{AUC of \translateClassifyLlama and \translateClassifyTowerBlocks using a fixed English classifier, \texttt{toxic-bert}, and a \classify pipeline using a fixed multilingual classifier, \texttt{distilbert-base-multilingual-cased-toxicity}. \textbf{Using a finetuned LLM for translation improves pipeline performance for higher-resourced languages.}}
    \label{fig:translation-benefit-by-language-instruct-vs-finetuned}
\end{figure*}

\begin{figure}[h]
    \centering
    \includegraphics[height=5cm]{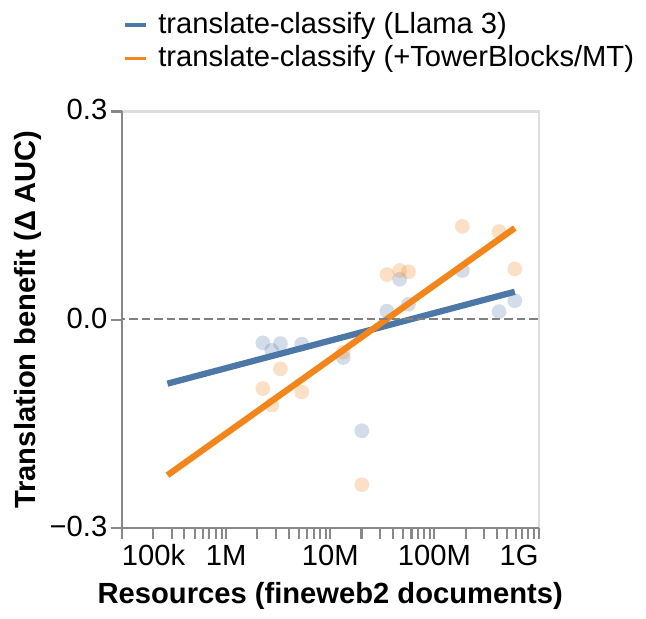}
    \vspace{6pt}
    \caption{Change in AUC (i.e., translation benefit) of \translateClassifyLlama and \translateClassifyTowerBlocks pipelines with a fixed English classifier against a fixed \classify pipeline with a fixed multilingual classifier, as a function of language resources. \textbf{The \translateClassifyTowerBlocks pipeline performance is more sensitive to available language resources.}}
    \label{fig:translation-benefit-vs-resources-llama-vs-mt-finetuned}
\end{figure}

\begin{figure*}[!t]
    \centering
    \includegraphics[width=\linewidth,trim={0.25cm 0 0.25cm 0},clip]{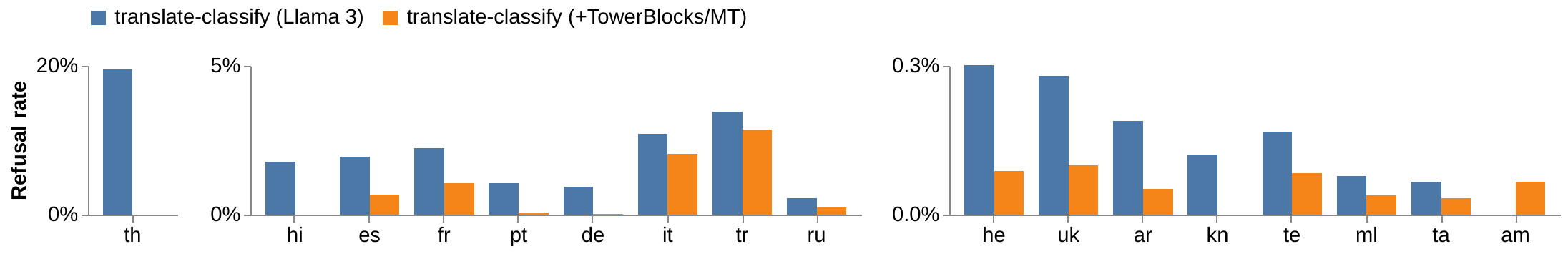}
    \caption{Translation refusal rate of \translateClassifyLlama and \translateClassifyTowerBlocks pipelines. Note three separate scales for legibility. \textbf{Using a finetuned LLM for translation reduces refusal rates.}}
    \label{fig:effect-refusal}
\end{figure*}

\begin{table*}[t]
\centering
\small
\setlength{\tabcolsep}{4pt}
\begin{tabular}{lccc}
\toprule
Name & TPR (\texttt{Llama 3.1}) & TNR (\texttt{+TowerBlocks/MT}) \\
\midrule
TH & 100\% & 100\% \\
DE & 100\% & 100\% \\
UK & 70\% & 100\% \\
ML & 75\% & 100\% \\
AR & 40\% & 100\% \\
\bottomrule
\end{tabular}
\caption{Analysis of human annotations of refusal predictions, showing True Positive Rate (TPR) of Llama 3.1 8B Instruct (\texttt{Llama 3.1}) and the True Negative Rate (TNR) of the same model finetuned on TowerBlocks/MT (\texttt{+TowerBlocks/MT}). \textbf{Refusal detection is highly accurate for Thai, German, and Ukrainian. Even for Malayalam and Levantine Arabic where the refusal detector produces some false positives, the finetuned model never refuses.}}
\label{tab:refusal_mitigation}
\end{table*}

When using safety-tuned LLMs for translation, a we noticed that key risk is \emph{refusal}: the model declines to translate inputs containing harmful or toxic content, which can severely limit coverage in toxicity detection.  
We examine whether finetuning for MT can mitigate this problem by comparing two \translateClassify pipelines:  
(1) \translateClassifyLlama, which uses translations from a standard instruction-tuned LLM (Llama 3.1 8B Instruct), and  
(2) \translateClassifyTowerBlocks, which uses translations from the same base model after supervised finetuning (MT-SFT) on the TowerBlocks/MT dataset \citep{tower_llm_2024} (see \cref{sec:app-mt-finetuning} for details).  
Both pipelines feed translations to a fixed English-only classifier, \texttt{toxic-bert}, to isolate translation effects, and are compared against a direct multilingual \classify pipeline using \texttt{distilbert-base-multilingual-cased -toxicity}.

\textbf{Refusal detection} \quad
We use \texttt{Minos} \citep{minos} to assign each translation output $y_i = T(x_i)$ a refusal probability $P_r(y_i)$.  
The refusal rate is defined as:
\begin{align*}
    \textrm{R}(T) = \frac{1}{N} \sum_{i=1}^{N} [P_r(T(x_i)) > 0.95] ,
\end{align*}
where a $0.95$ threshold minimizes false positives.  
For two systems $T_A$ and $T_B$, the difference in refusal rates is:
\begin{align*}
    \Delta \textrm{R}(P_{T_A}, P_{T_B}) = \textrm{R}(P_{T_A}) - \textrm{R}(P_{T_B}).
\end{align*}

\begin{figure*}[!t]
    \centering
    \begin{subfigure}[t]{0.4\textwidth}
        \centering 
        \includegraphics[height=5cm]{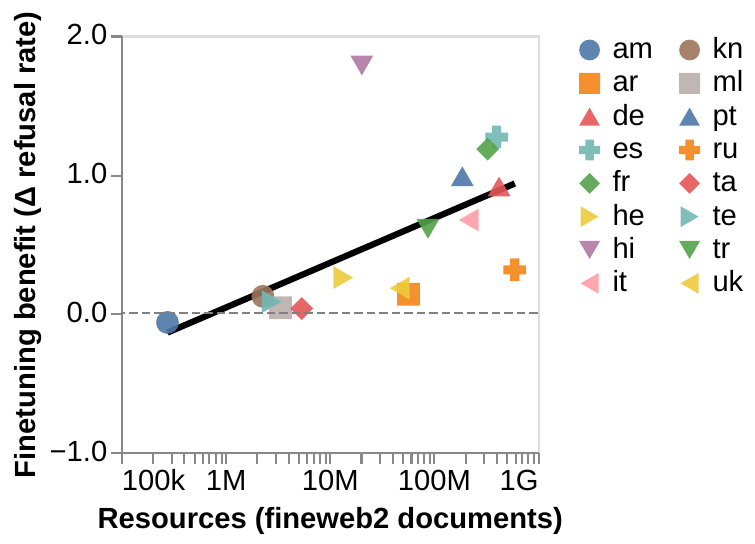}
        \caption{$\Delta$ refusal}
        \label{fig:finetuning-benefit-refusal-vs-resources}
    \end{subfigure}
    \hspace{20pt}
    \begin{subfigure}[t]{0.4\textwidth}
        \centering 
        \includegraphics[height=5cm]{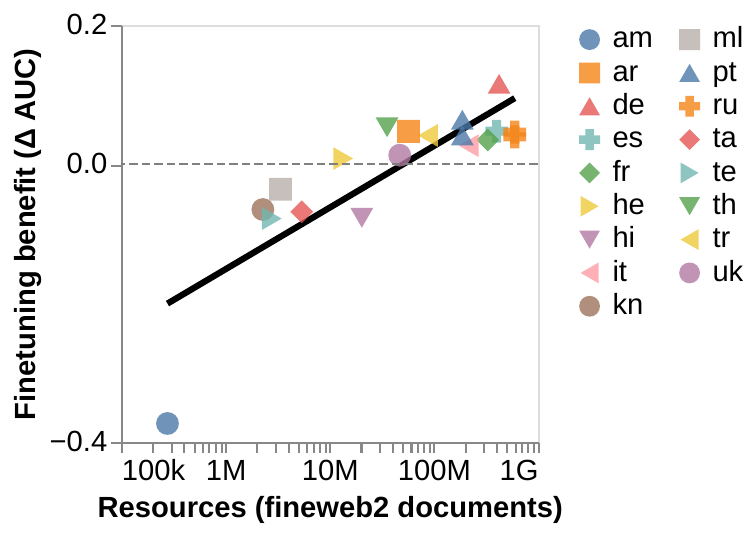}
        \caption{$\Delta$ AUC}
        \label{fig:finetuning-benefit-vs-resources}
    \end{subfigure}
    \caption{Change in \textbf{(a)} translation refusal rate and \textbf{(b)} AUC  of a \translateClassifyTowerBlocks pipeline against a \translateClassifyLlama pipeline, both with a fixed English classifier, \texttt{toxic-bert}, as a function of language resources. \textbf{The benefit of using a finetuned LLM for translation, in terms of both refusal rates and improved performance, increases for with language resources.}}
    \label{fig:finetuning-benefit}
\end{figure*}

\textbf{Refusal results} \quad
As shown in \cref{fig:effect-refusal}, \translateClassifyTowerBlocks reduces refusal rates in \emph{every} language compared with \translateClassifyLlama.  
The reduction scales approximately log-linearly with language resources (\cref{fig:finetuning-benefit-refusal-vs-resources}), indicating that MT-SFT particularly benefits high-resource languages where refusals are rarer but still impactful.  
Lower refusal means more toxic content is actually processed by the classifier, directly improving pipeline coverage.

\textbf{Human verification of refusal mitigation} \quad
To validate both the accuracy of our automated refusal detection and the effectiveness of MT-SFT in addressing refusals, we conducted a targeted human annotation study. For each dataset, we randomly sampled up to 5\% of the content flagged as refusals by the base Llama 3.1 8B Instruct model, with a minimum of 10 examples per dataset. Annotators manually verified whether each flagged case was indeed a refusal, then examined translations of the same inputs generated by the MT-finetuned model. As shown in \cref{tab:refusal_mitigation}, the refusal detector achieved perfect true positive rates for Thai, German, and Ukrainian, and high —though not perfect— accuracy for Malayalam and Levantine Arabic, where some false positives were observed. Importantly, the MT-finetuned model produced valid translations for \emph{all} annotated examples, yielding a true negative rate of 100\% across every language in the sample. This confirms that, at least for the languages tested, MT-SFT can completely eliminate refusals observed in the base instruction-tuned model, turning previously blocked content into usable inputs for the downstream classifier.

\textbf{MT-SFT improves performance for high resource languages} \quad
In addition to lowering refusals, MT-SFT also improves classification accuracy.  
In \cref{fig:translation-benefit-by-language-instruct-vs-finetuned}, \translateClassifyTowerBlocks achieves higher AUC than \translateClassifyLlama for 11 of 17 languages, with gains concentrated in high-resource settings.  
When measured against the multilingual \classify baseline, \translateClassifyTowerBlocks shows even stronger sensitivity to language resource availability (\cref{fig:translation-benefit-vs-resources-llama-vs-mt-finetuned}).  

\subsection{LLM judges underperform on lower-resourced languages}

\label{sec:results-llm-judge}

Given the strong performance of LLMs across a range of tasks, we additionally compare pipelines based on traditional classifiers vs. zero-shot LLM judges.

Specifically, we analyze the performance of two judge-based pipelines: (1) \judge, where an LLM is prompted to classify untranslated samples, and (2) \translateJudge, where the LLM is prompted to classify samples that have been translated into English.
The judge in both cases is Llama 3.1 8B Instruct with the following custom prompt, where \texttt{\{\{sample\}\}} is either the translated or original sample according to pipeline: \hfill \newline

\begin{lstlisting}
You are a toxicity classifier. Classify the following text as toxic or not toxic. Respond with 'toxic' or 'not toxic' with no other comments or explanation.
Text: {{sample}}
Classification:
\end{lstlisting}

Translations are also produced using Llama 3.1 8B Instruct.
We compare both judge pipelines against a \classify pipeline based on the overall best English classifier, \texttt{toxic-bert}, coupled again with Llama 3.1 8B Instruct for fair comparison. 
We evaluate all pipelines using accuracy as opposed to AUC, as the LLM judge emits only a label, rather than a continuous score. 

\begin{figure*}[!t]
    \centering
    \includegraphics[width=\linewidth,trim={0.25cm 0 0.25cm 0},clip]{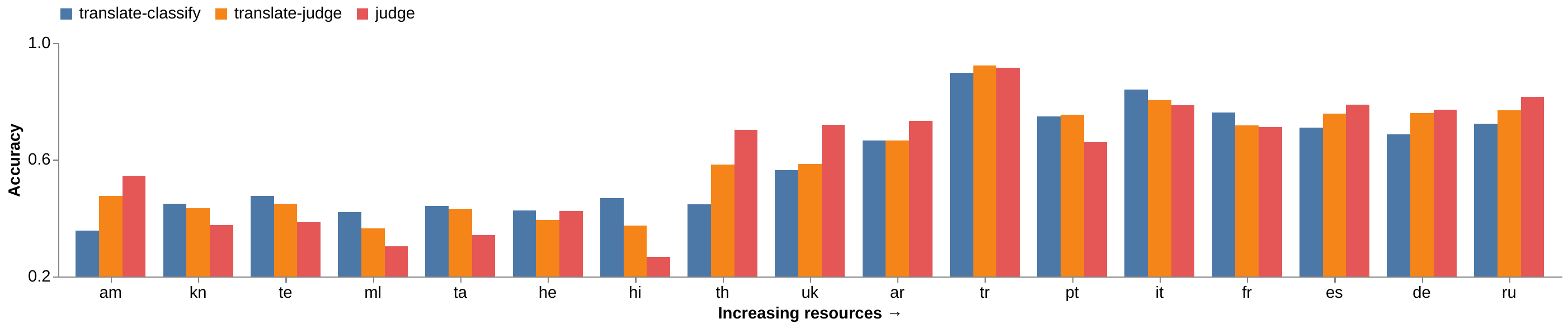}
    \caption{Accuracy of \translateJudge, \judge, and \translateClassify with a fixed English classifier, \texttt{toxic-bert}, all using a Llama 3 for translation. \textbf{Translation with traditional classifiers outperforms LLM judges for most lower resourced languages.}}
    \label{fig:llm-judge}
\end{figure*}

\textbf{Results} \quad
\Cref{fig:llm-judge} illustrates notable patterns in the comparative performance of \translateJudge and \judge pipelines.
Across all languages, translation-based approaches narrowly outperform their untranslated counterparts; however, this advantage becomes pronounced in low-resource settings, where \translateJudge completely dominates, outperforming \judge in 6 out of 7 low-resource languages.
Similarly, \translateClassify pipelines provide a slight overall edge compared to both \judge and \translateJudge, but the margin is especially significant for low-resource languages, where \translateClassify overwhelmingly wins (again in 6 out of 7 cases). 
These results further indicate that multilingual capabilities in LLMs are not homogeneously distributed: while MT models demonstrate broader multilingual reach, toxicity classification performance by LLMs is markedly less consistent across lower-resource languages.

\section{Discussion}
\label{sec:discussion}

Across ten benchmarks spanning 17 languages, our analysis suggests that translation-based approaches can be successfully leveraged to support multilingual toxicity detection at scale.
Specifically, we observe that \translateClassify pipelines outperform \classifyOOD, a non-finetuned classifier operating OOD (i.e., an off-the-shelf model) in the majority of cases, and can even occasionally outperform \classifyID, dedicated finetuned classifiers evaluated ID. 
The relative benefit of using \translateClassify over \classify pipelines increases with both a language's available resources and the quality of the translation system. 
This may suggest that while translation may be an effective strategy \emph{in general}, it does have the potential to increase performance disparities between better- and worse-resourced languages.
We additionally note that using an MT-finetuned LLM for translations can further drive up pipeline performance, in part, by reducing refusal rates, but that this benefit appears to be reserved for higher-resourced languages. 
Finally, we evaluate the utility of an LLM judge approach over traditional (e.g., BERT-based) classification, finding that in lower-resourced languages, \translateClassify consistently outperforms. 

\textbf{Practical recommendations} \quad
We make four practical recommendations for practitioners looking to deploy multilingual toxicity detection at scale.
\begin{enumerate}
    \item At the very least, \translateClassify pipelines using traditional classifiers and LLM-based translation should be considered a robust baseline.
    \item If fine-tuning on dedicated data is unavailable, a \translateClassify pipeline is likely to provide a strong first choice of model, particularly in languages where translation quality is high. 
    \item If operating on a higher-resourced language, making use of an MT-finetuned LLM may offer some performance improvements over a standard instruction-tuned LLM, particularly in the scenario where refusal rates can be reduced.
    \item Unlike many other NLP tasks, an LLM judge demonstrates only a limited performance advantage on select higher-resourced languages when compared to traditional (e.g., BERT-based) classifiers.
    \end{enumerate}

\section*{Limitations}
While we approach multilingual toxicity detection through the lens of a practitioner making a choice between available, off-the-shelf pipeline components, this does limit our ability to analyze the role of specific finetuning details. 
For example, in contrast with previous work \citep{artetxe-etal-2023-revisiting} that has contrasted cross-lingual transfer pipelines where the classifier was finetuned on either the original domain or the outputs of the translation system, we only make use of publicly-available classifiers which may be finetuned on different numbers of samples or different domains, and none of which are finetuned on translations.
However, given the performance improvements offered by the \translateClassify pipeline \emph{without finetuning on translations}, we might expect a translation-finetuned classifier to further benefit the \translateClassify approach. 

As we note in \cref{sec:methods-data}, our work is also potentially limited by shifts in data distribution between languages. 
In order to identify broad trends across many languages with different levels of resources, we draw samples from different constituent datasets. 
These datasets, however, are drawn from different domains (e.g., social media vs. WikiMedia talk pages) with labels produced using different annotation schemas (e.g., identifying hate speech vs. toxicity). 
As a result, our conclusions should be interpreted as indicative of general trends about the relative utility of translation, rather than individual claims about how well translation may function on any given language.
This limitation could be overcome with access to additional highly-multilingual datasets of labeled toxicity data.

\FloatBarrier
\bibliography{anthology,custom,bib-tox}
\bibliographystyle{assets/plainnat}

\clearpage
\newpage

\appendix

\setcounter{figure}{0}
\renewcommand{\figurename}{Fig.}
\renewcommand{\thefigure}{S\arabic{figure}}

\setcounter{table}{0}
\renewcommand{\tablename}{Table}
\renewcommand{\thetable}{S\arabic{table}}

\section{Additional methods}

\subsection{Toxicity classifier selection}
\label{sec:app-toxicity-classifier-selection}

We evaluate on a sample of toxicity classifiers that are publicly-available on Hugging Face. 
We reviewed classifiers that matched the search terms ``toxic'' and ``toxicity'', selecting those that supported
either English or one or more of the 17 languages analyzed. 
Classifiers were limited to those that were permissively-licensed, with clear data provenance (to allow for distinguishing between ID and OOD performance), and substantial community engagement (as measured by downloads and likes).
See \cref{table:models} for all classifiers evaluated.

\subsection{MT finetuning an LLM}

We used Llama 3.1 8b Instruct as our baseline model and finetuned it for 5 epochs with the 
MT split from Towerblocks 0.2, a multi-task, multilingual SFT  dataset. We employed the AdamW optimizer with a learning rate initialized to \(1 \times 10^{-6}\), \(\beta_1\) and \(\beta_2\) coefficients set to 0.9 and 0.95 respectively, and a weight decay of 0.1. We used a cosine annealing learning rate scheduler configured with a final learning rate scaled to 0.2 times the initial rate and a total of 1,000 warmup steps.

\label{sec:app-mt-finetuning}

\section{Additional results}

In \cref{table:overall-results} we present the detailed results behind \cref{fig:absolute-performance-best-possible-vs-ood,fig:absolute-performance-best-possible-vs-id}, showing the performance of the best-possible \translateClassify, \classifyOOD, and \classifyID pipelines over all languages.
In \cref{table:best-pipeline-translated,table:best-pipeline-ood,table:best-pipeline-id} we present the corresponding best-performing translation system and classifier combinations for \translateClassify, \classifyOOD, and \classifyID respectively.

\begin{figure*}[!th]
    \centering
    \includegraphics[width=\linewidth]{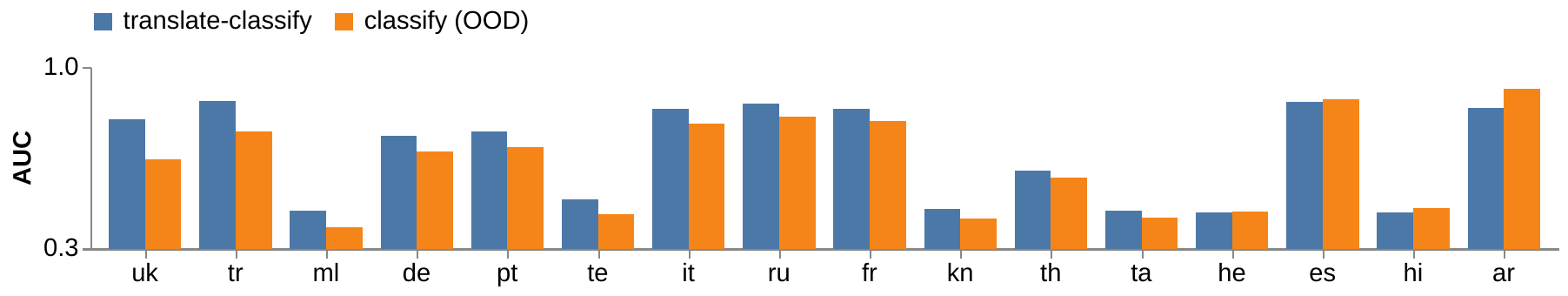}
    \caption{Translation-based toxicity detection pipelines with a fixed English-supporting classifier, \texttt{distilbert-base-multilingual-cased-toxicity}, outperform off-the-shelf pipelines across 12 out of 16 evaluated languages.}
    \label{fig:absolute-performance-best-possible-fixed-classifier-vs-ood}
\end{figure*}

In \cref{fig:absolute-performance-best-possible-fixed-classifier-vs-ood}, we present a version of \cref{fig:absolute-performance-best-possible-vs-ood} but reducing one degree of freedom: rather than choosing the best-possible combination of translation system and classifier, here we choose the best possible translation system though use a fixed classifier, \texttt{distilbert-base-multilingual-cased-toxicity}.
In this setting, \translateClassify still outperforms across 12 of 16 languages.

\begin{table*}[h]
\footnotesize
\centering
\begin{tabular}{lrrr}
\toprule
& \multicolumn{3}{c}{\textbf{AUC}} \\
Language & ID & OOD & Translated \\
\midrule
ar & - & \bfseries 0.92 & 0.89 \\
he & - & \bfseries 0.44 & 0.44 \\
hi & - & \bfseries 0.46 & 0.44 \\
kn & - & 0.42 & \bfseries 0.45 \\
ml & - & 0.38 & \bfseries 0.45 \\
pt & - & 0.69 & \bfseries 0.79 \\
ta & - & 0.42 & \bfseries 0.44 \\
te & - & 0.43 & \bfseries 0.49 \\
th & - & 0.57 & \bfseries 0.67 \\
uk & - & 0.64 & \bfseries 0.85 \\
am & 0.88 & - & \bfseries 0.99 \\
de & 0.81 & 0.67 & \bfseries 0.82 \\
es & \bfseries 0.92 & 0.88 & 0.91 \\
fr & \bfseries 0.88 & 0.79 & 0.88 \\
it & \bfseries 0.88 & 0.78 & 0.88 \\
ru & \bfseries 0.97 & 0.81 & 0.90 \\
tr & 0.94 & 0.75 & \bfseries 0.96 \\
\bottomrule
\end{tabular}
\caption{Best possible performance over all languages. Where a finetuned classifier isn't available, translation-based pipelines often outperform. }
\label{table:overall-results}
\end{table*}

\begin{table*}[p]
\footnotesize
\centering
\begin{tabular}{llr}
\toprule
Language & Classifier & AUC \\
\midrule
am & textdetox/xlmr-large-toxicity-classifier & 0.88 \\
de & ml6team/distilbert-base-german-cased-toxic-comments & 0.81 \\
es & unitary/multilingual-toxic-xlm-roberta & 0.92 \\
fr & unitary/multilingual-toxic-xlm-roberta & 0.88 \\
it & unitary/multilingual-toxic-xlm-roberta & 0.88 \\
ru & s-nlp/russian\_toxicity\_classifier & 0.97 \\
tr & unitary/multilingual-toxic-xlm-roberta & 0.94 \\
\bottomrule
\end{tabular}
\caption{Best-performing ID pipeline per language.}
\label{table:best-pipeline-id}
\end{table*}

\begin{table*}[p]
\footnotesize
\centering
\begin{tabular}{llr}
\toprule
Language & Classifier & AUC \\
\midrule
ar & textdetox/xlmr-large-toxicity-classifier & 0.92 \\
de & citizenlab/distilbert-base-multilingual-cased-toxicity & 0.67 \\
es & textdetox/xlmr-large-toxicity-classifier & 0.88 \\
fr & citizenlab/distilbert-base-multilingual-cased-toxicity & 0.79 \\
he & citizenlab/distilbert-base-multilingual-cased-toxicity & 0.44 \\
hi & citizenlab/distilbert-base-multilingual-cased-toxicity & 0.46 \\
it & citizenlab/distilbert-base-multilingual-cased-toxicity & 0.78 \\
kn & citizenlab/distilbert-base-multilingual-cased-toxicity & 0.42 \\
ml & citizenlab/distilbert-base-multilingual-cased-toxicity & 0.38 \\
pt & unitary/multilingual-toxic-xlm-roberta & 0.69 \\
ru & citizenlab/distilbert-base-multilingual-cased-toxicity & 0.81 \\
ta & citizenlab/distilbert-base-multilingual-cased-toxicity & 0.42 \\
te & citizenlab/distilbert-base-multilingual-cased-toxicity & 0.43 \\
th & citizenlab/distilbert-base-multilingual-cased-toxicity & 0.57 \\
tr & citizenlab/distilbert-base-multilingual-cased-toxicity & 0.75 \\
uk & citizenlab/distilbert-base-multilingual-cased-toxicity & 0.64 \\
\bottomrule
\end{tabular}
\caption{Best-performing OOD pipeline per language.}
\label{table:best-pipeline-ood}
\end{table*}

\begin{table*}[p]
\footnotesize
\centering
\begin{tabular}{lllr}
\toprule
Language & Translation system & Classifier & AUC \\
\midrule
am & Llama 3.1 8B Instruct & unitary/toxic-bert & 0.99 \\
ar & GPT-4o & unitary/toxic-bert & 0.89 \\
de & GPT-4o & unitary/multilingual-toxic-xlm-roberta & 0.82 \\
es & GPT-4o & unitary/toxic-bert & 0.91 \\
fr & GPT-4o & unitary/toxic-bert & 0.88 \\
he & Llama 3.1 8B TowerBlocks & citizenlab/distilbert-base-multilingual-cased-toxicity & 0.44 \\
hi & Llama 3.1 8B Instruct & citizenlab/distilbert-base-multilingual-cased-toxicity & 0.44 \\
it & GPT-4o & unitary/toxic-bert & 0.88 \\
kn & Llama 3.1 8B Instruct & citizenlab/distilbert-base-multilingual-cased-toxicity & 0.45 \\
ml & Llama 3.1 8B Instruct & citizenlab/distilbert-base-multilingual-cased-toxicity & 0.45 \\
pt & GPT-4o & unitary/toxic-bert & 0.79 \\
ru & GPT-4o & unitary/multilingual-toxic-xlm-roberta & 0.90 \\
ta & Llama 3.1 8B Instruct & citizenlab/distilbert-base-multilingual-cased-toxicity & 0.44 \\
te & Llama 3.1 8B Instruct & citizenlab/distilbert-base-multilingual-cased-toxicity & 0.49 \\
th & GPT-4o & textdetox/xlmr-large-toxicity-classifier & 0.67 \\
tr & Llama 3.1 8B TowerBlocks & unitary/toxic-bert & 0.96 \\
uk & GPT-4o & unitary/multilingual-toxic-xlm-roberta & 0.85 \\
\bottomrule
\end{tabular}
\caption{Best-performing translated pipeline per language.}
\label{table:best-pipeline-translated}
\end{table*}

\end{document}